\documentclass[journal, letterpaper]{IEEEtran}
\usepackage{graphicx}
\usepackage{url}
\usepackage{cite}
\usepackage{amsmath}
\usepackage{amssymb}
\usepackage{textgreek}
\usepackage{listings}
\usepackage{csvsimple}
\usepackage{longtable}
\usepackage{color}
\usepackage{booktabs}

\usepackage{microtype}
\usepackage{hyperref}
\usepackage{mathtools}
\usepackage{amsthm}
\usepackage[capitalize,noabbrev]{cleveref}

\theoremstyle{plain}
\newtheorem{theorem}{Theorem}[section]
\newtheorem{proposition}[theorem]{Proposition}

\theoremstyle{definition}

\theoremstyle{remark}

\title{Preserving Continuous Symmetry in Discrete Spaces: Geometric-Aware Quantization for SO(3)-Equivariant GNNs}

\author{Haoyu Zhou$^{1}$, Ping Xue$^{2}$, Hao Zhang$^{2}$, Tianfan Fu$^{1}$%
\thanks{$^{1}$Nanjing University, Nanjing, China.}%
\thanks{$^{2}$Gusu Lab, Suzhou, China.}%
}

\begin{document}
\markboth{Course Name/Number: Project Name}{}
\maketitle

\begin{abstract}
Equivariant Graph Neural Networks (GNNs) are essential for physically consistent molecular simulations but suffer from high computational costs and memory bottlenecks, especially with high-order representations. While low-bit quantization offers a solution, applying it naively to rotation-sensitive features destroys the $SO(3)$-equivariant structure, leading to significant errors and violations of conservation laws. To address this issue, in this work, we propose a \textbf{Geometric-Aware Quantization (GAQ)} framework that compresses and accelerates equivariant models while rigorously preserving continuous symmetry in discrete spaces. Our approach introduces three key contributions: (1) a Magnitude–Direction Decoupled Quantization (MDDQ) scheme that separates invariant lengths from equivariant orientations to maintain geometric fidelity; (2) a symmetry-aware training strategy that treats scalar and vector features with distinct quantization schedules; and (3) a robust attention normalization mechanism to stabilize gradients in low-bit regimes. Experiments on the rMD17 benchmark demonstrate that our W4A8 models match the accuracy of FP32 baselines (9.31 meV vs. 23.20 meV) while reducing Local Equivariance Error (LEE) by over $30\times$ compared to naive quantization. On consumer hardware, GAQ achieves $2.39\times$ inference speedup and $4\times$ memory reduction, enabling stable, energy-conserving molecular dynamics simulations for nanosecond timescales.
\end{abstract}

\begin{figure*}[t]
  \centering
  \includegraphics[width=\linewidth]{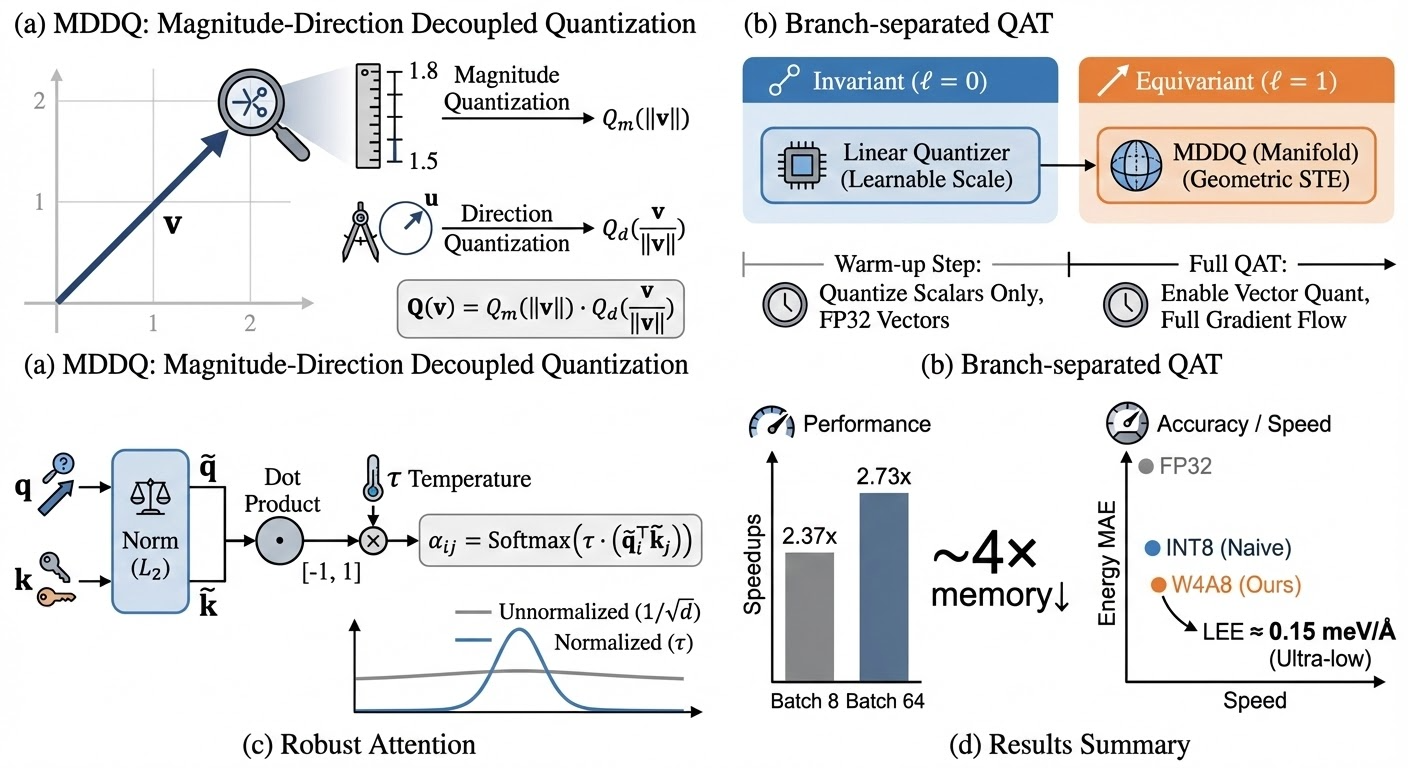}
  \caption{\textbf{Overview of the proposed equivariant quantization framework.} 
  (a) \textbf{MDDQ}: Decouples equivariant vectors into magnitude $\|\mathbf{v}\|$ and direction $\mathbf{u}$ to preserve geometric orientation under low precision. 
  (b) \textbf{Branch-separated QAT}: Treats invariant and equivariant features differently with a staged training schedule. 
  (c) \textbf{Robust Attention}: Stabilizes dot-products via $\ell_2$ normalization and temperature scaling $\tau$. 
  (d) \textbf{Results Summary}: Our method achieves $2.37$--$2.73\times$ faster inference and $\sim 4\times$ memory reduction with ultra-low equivariance error ($\text{LEE} \approx 0.15$).}
  \label{fig:teaser}
\end{figure*}
\section{Introduction}

Many important problems in science and engineering involve data with inherent geometric symmetries.  In particular, 3D $SO(3)$-equivariant GNNs—neural networks whose outputs transform consistently under 3D rotations—have become pivotal for molecular property prediction and simulation.  Recent $SO(3)$-equivariant networks like NequIP and So3krates achieve state-of-the-art accuracy on small molecules and periodic systems by explicitly enforcing rotational symmetry~\cite{batzner20223,frank2022so3krates}.  By encoding physical rotation symmetry, these models honor fundamental conservation laws: according to Noether’s theorem, each continuous symmetry corresponds to a conserved quantity, so rotational invariance implies conservation of angular momentum~\cite{noether1971invariant}.  Early work on energy-conserving force fields leveraged equivariant neural networks to guarantee energy conservation~\cite{chmiela2017machine}.  Enforcing $SO(3)$ equivariance in learned interatomic potentials thus helps ensure that predictions remain physically consistent under arbitrary orientation changes.  In MD simulations, such symmetry preservation is critical; if a model’s rotational equivariance is broken, it can introduce spurious torques or artifacts, leading to violations of conservation (manifesting, for instance, as a gradual energy drift over long trajectories)~\cite{muller2023exact,musaelian2023learning}.

Despite their superior accuracy and physical faithfulness, equivariant GNNs are computationally expensive.  Their expressiveness often relies on high-order geometric tensor products and elaborate basis representations, causing the number of operations and parameters to grow combinatorially with model depth and order.  This can lead to a combinatorial explosion in both computation and memory footprint, especially as one incorporates higher-order $SO(3)$ features.  Earlier works have noted that the cost of these representations can hinder scalability, and have even sought to eliminate expensive tensor-product operations to improve efficiency~\cite{thomas2018tensor,fuchs2020se,batatia2022mace}.  Recent efforts to construct equivariant multilayer perceptrons or to identify optimal invariant bases~\cite{finzi2021practical,allen2025optimalinvariantbasesatomistic,meng2024towards} further underscore the tension between expressivity and efficiency in equivariant models.  In addition, modern high-performance computing faces the memory wall: memory bandwidth and capacity have not kept pace with arithmetic throughput.  Large-scale or long-timescale simulations may require millions of GNN evaluations, making memory access the primary bottleneck.  As a result, full-precision (32-bit) equivariant models become impractical for such scenarios due to excessive memory traffic and slow inference.  Innovations such as FlashAttention and other I/O-aware kernels for transformers~\cite{dao2022flashattention,dehghani2023scaling} and more efficient Euclidean transformers for force-field prediction~\cite{frank2024euclidean} have been proposed to reduce memory overhead, yet the challenge remains particularly acute in $SO(3)$-equivariant GNNs, where each hidden channel carries multiple irreducible components, compounding memory and compute demands.

Low-bit quantization is a natural solution to these efficiency challenges.  By using reduced numerical precision (e.g., 8-bit integers), we can shrink model size and accelerate memory-bound operations, effectively mitigating the memory wall.  The deep-learning community has explored quantization extensively: early low-bit convolutional networks like DoReFa-Net~\cite{zhou2016dorefa}, LSQ~\cite{esser2019learned}, and quantization-aware training for vision models~\cite{jacob2018quantization} paved the way; more recent techniques such as QDrop~\cite{wei2022qdrop}, SmoothQuant~\cite{xiao2023smoothquant} and ZeroQuant~\cite{yao2022zeroquant} provide high-quality post-training quantization for large language models.  In graph settings, Degree-Quant extends quantization-aware training to graph neural networks~\cite{tailor2020degree}.  However, these generic approaches are \emph{geometry-agnostic}: they treat feature channels as unstructured scalars.  When applied to equivariant GNNs, naive quantization of vector components ($\ell=1$) on Cartesian axes destroys the algebraic relationships required by Wigner-$D$ matrices, leading to \emph{quantization-induced symmetry breaking}.  Naively discretizing continuous-valued features on a fixed grid introduces small invariant errors that accumulate through network layers.  In an $SO(3)$-equivariant GNN, even minor rounding inconsistencies can violate equivariance.  This symmetry breaking is especially problematic in physics-driven applications: if rotational symmetry is not rigorously preserved, the model may produce outputs that drift from true physics over time (e.g., predicting forces that do not conserve energy or angular momentum).  The central question is how to preserve continuous $SO(3)$ symmetry within a discrete, quantized neural network.

In this work, we propose a geometric-aware quantization framework that resolves the above conflict by design.  The core idea is to incorporate group-theoretic structure into the quantization process, so that discrete representations remain as equivariant as their continuous counterparts.  We build upon the insight that a 3D vector can be factored into an invariant magnitude and an equivariant direction on the unit sphere, following the representation theory of $SO(3)$ used in tensor field networks and SE(3) transformers~\cite{thomas2018tensor,fuchs2020se}.  By quantizing these components separately (and judiciously), we can reduce precision while controlling the error in orientations.  Specifically, we introduce a magnitude–direction decoupled quantization (MDDQ) scheme that quantizes the norm of each vector and the direction (unit vector) on $S^2$ independently.  This decoupling ensures that rotations of the input primarily affect the directional part, which is handled on a discrete spherical codebook designed to respect $SO(3)$ rotations.  We show that if the directional quantizer commutes with rotations, then the overall quantization is equivariant up to bounded error.

In addition to the quantization scheme itself, maintaining accuracy under low precision requires careful training strategies and architectural modifications.  We propose a symmetry-aware quantization-aware training approach in which we branch-separate the network’s channels by their transformation properties.  In our $SO(3)$-equivariant transformer architecture, some feature channels are invariant scalars (e.g., chemical attributes or learned radial features) while others are equivariant vectors or tensors (geometric features).  We treat these two types of features differently during quantization and training: invariant channels can be quantized more aggressively or with different calibration than equivariant channels.  Conversely, equivariant feature channels are quantized in a manner that prioritizes preserving their transformation behavior.  This branch-separated training ensures that quantization noise does not disproportionately harm the equivariant components of the model.  We also incorporate a robust attention normalization mechanism designed for low-precision arithmetic.  The self-attention module, which mixes invariant and equivariant information, is modified to normalize its inputs (e.g., attention scores) in a way that counteracts the higher quantization error at low bit-widths.  Similar ideas have been explored for standard transformers in the context of query-key normalization~\cite{henry2020query} and scaling vision transformers~\cite{dehghani2023scaling}; here we tailor the normalization for $SO(3)$-equivariant data.

We evaluate our geometric-aware quantization framework on standard 3D molecular benchmarks (QM9 for molecular properties and rMD17 for molecular forces and energies).  Our 8-bit quantized $SO(3)$-equivariant GNN achieves comparable accuracy to full 32-bit models on energy and force prediction tasks, indicating that our symmetry-preserving quantization does not degrade predictive performance.  Just as importantly, we measure the equivariance error of the network using the Local Equivariance Error (LEE) metric and find that our quantized models retain near-perfect equivariance, in stark contrast to naive quantization approaches that exhibit significant symmetry error.  Empirically, the quantized models remain stable in long MD simulations, showing no noticeable energy drift or other signs of physical law violation.  Thanks to the $4\times$ reduction in model memory and optimized low-bit operations, we observe $2.37$–$2.73\times$ faster inference per molecule on a CPU, enabling faster simulation rates.  These gains underscore that quantization can be a mathematically principled tool (not just a brute-force compression) to overcome the combinatorial complexity of equivariant models and the memory wall of modern hardware.

\noindent (1) \textbf{Geometric-aware quantization via MDDQ:} We introduce a magnitude–direction decoupled quantization scheme that quantizes each vector’s length and direction separately.  This group-theoretic approach preserves $SO(3)$ equivariance on the discrete quantization grid, preventing symmetry breaking even at 8-bit precision.

\noindent (2) \textbf{Symmetry-aware branch-separated QAT:} We develop a quantization-aware training strategy that splits the model into invariant and equivariant branches.  By handling scalar and geometric features with tailored quantization and calibration, we maintain high accuracy under low precision while rigorously respecting the different roles each channel plays in preserving symmetry.

\noindent (3) \textbf{Equivariant attention normalization:} We propose an $SO(3)$-equivariant attention normalization mechanism that stabilizes the transformer’s attention calculations in low-bit settings.  This technique improves the robustness of attention operations against quantization noise, drawing inspiration from query-key normalization~\cite{henry2020query,dehghani2023scaling}, and further ensures that the overall network remains stable and symmetry-preserving after quantization.

\begin{figure}[t]
  \centering
  \includegraphics[width=\linewidth]{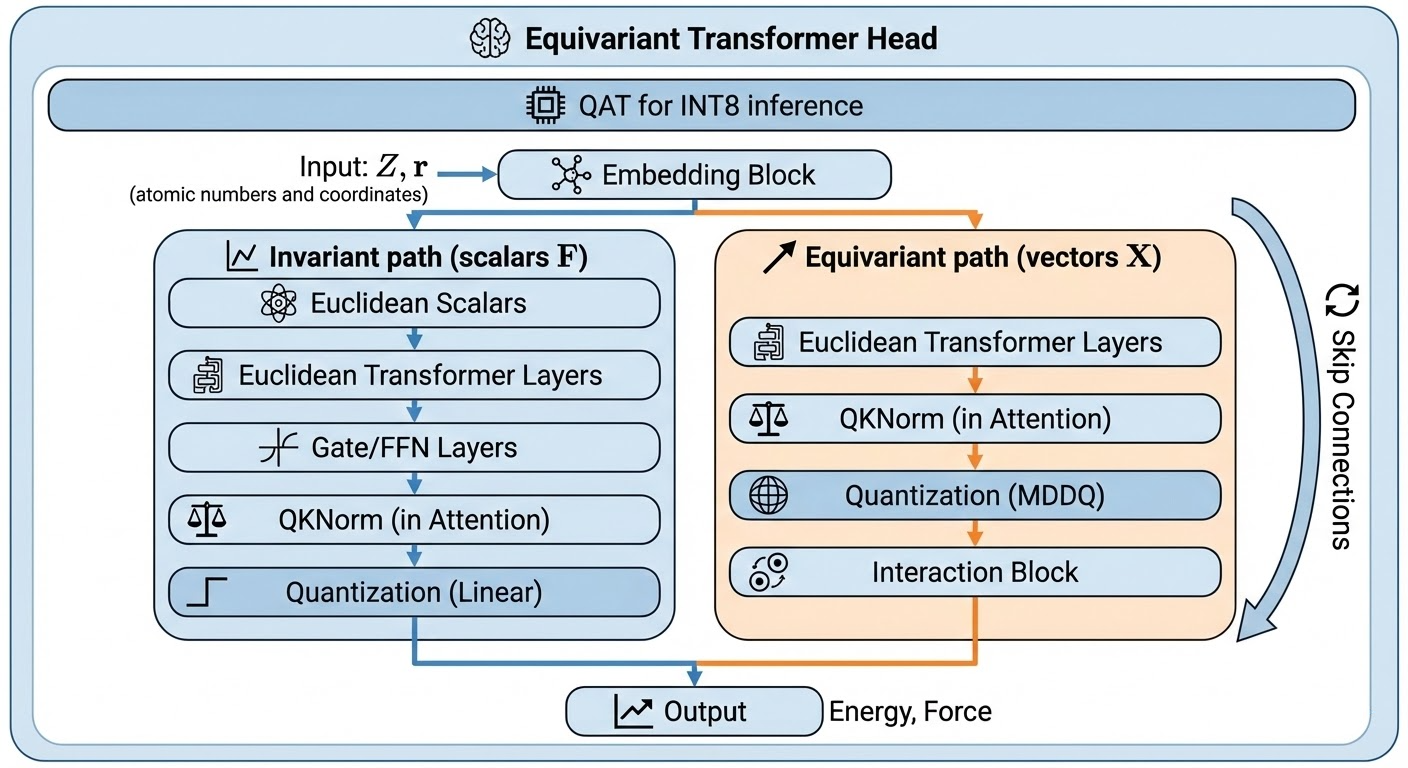}
  \caption{\textbf{Detailed architecture and quantization pipeline.} Inputs $Z$ and $\mathbf{r}$ are processed through decoupled invariant and equivariant paths, employing specialized quantization strategies (Linear vs. MDDQ) for each branch.}
  \label{fig:arch_pipeline}
\end{figure}

\section{Related Work}

\noindent\textbf{Geometric Deep Learning and Equivariance.}
Incorporating physical symmetries into neural networks is foundational for molecular modeling.  Early works established group convolutions for discrete symmetries~\cite{cohen2016group} and continuous symmetries such as rotations and reflections via harmonic networks and steerable filters~\cite{worrall2017harmonic,weiler20183d}.  In 3D atomistic tasks, architectures like Tensor Field Networks~\cite{thomas2018tensor} and SE(3)-Transformers~\cite{fuchs2020se} enforce equivariance via spherical harmonics and tensor products.  Recent state-of-the-art models, such as NequIP~\cite{batzner20223} and So3krates~\cite{frank2022so3krates}, leverage higher-order irreps ($\ell \ge 2$) to capture complex many-body interactions.  However, the computational cost of mixing these high-$\ell$ features scales super-linearly~\cite{batatia2022mace}, creating a bottleneck for large-scale simulations.  Alternatives such as 3D steerable CNNs~\cite{weiler20183d}, harmonic networks~\cite{worrall2017harmonic} and recent Euclidean transformers~\cite{frank2024euclidean} explore different trade-offs between expressivity and efficiency.

\paragraph{Optimal Bases and Codebooks.}
Addressing the complexity of equivariant representations, Allen et al.~\cite{allen2025optimalinvariantbasesatomistic} theoretically characterized optimal invariant bases, suggesting that compact sets of invariants can losslessly represent atomic environments.  Our work operationalizes this insight for quantization: we construct a discrete \emph{spherical codebook} to approximate continuous orientations, effectively learning a minimal basis that balances efficiency with symmetry preservation.

\paragraph{Measuring Equivariance.}
Quantifying symmetry violation is critical for discretized models.  Gruver et al.~\cite{gruver2022lie} proposed a Lie derivative-based metric to measure sensitivity to infinitesimal rotations.  While valuable for global diagnosis, this continuous metric does not capture the discrete errors introduced by low-bit quantization.  In contrast, we utilize the \emph{Local Equivariance Error (LEE)}, which measures discrepancies under finite rotations and serves as a direct training signal to minimize quantization-induced symmetry breaking.

\noindent\textbf{Quantization in Geometric Contexts.}
Neural network quantization is a standard compression technique~\cite{jacob2018quantization}.  Advanced methods like LSQ~\cite{esser2019learned} and QDrop~\cite{wei2022qdrop} mitigate accuracy loss by learning step sizes or stochastic dropping.  However, these generic approaches are \emph{geometry-agnostic}: they treat feature channels as unstructured scalars.  When applied to equivariant GNNs, naive quantization of vector components ($\ell=1$) on Cartesian axes destroys the algebraic relationships required by Wigner-$D$ matrices, leading to \emph{quantization-induced symmetry breaking}. 

Prior GNN-specific works, such as Degree-Quant~\cite{tailor2020degree}, adapt quantization based on graph topology (e.g., node degree) but overlook geometric topology.  They fail to preserve the directionality of vector features.  Our work bridges this gap by proposing \textbf{Magnitude–Direction Decoupled Quantization (MDDQ)}.  Unlike prior arts, we perform quantization on the spherical manifold $S^2$, decoupling the invariant magnitude from the equivariant direction.  This ensures that the discrete representation respects the underlying $SO(3)$ geometry, enabling efficient deployment without sacrificing physical validity.

\section{Method}

We propose an $SO(3)$-equivariant quantization framework for transformer-based GNNs, aiming for fast low-bit inference without sacrificing accuracy or symmetry.  Our method builds on a So3krates-like architecture with separate invariant (scalar) and equivariant (vector) branches.  Before detailing the model, we briefly review the mathematical foundations of $SO(3)$-equivariance.  We then describe the architecture and our key contributions: a spectral codebook quantization for equivariant features, an optimization strategy on the unit-sphere manifold, a robust attention normalization, and an equivariance-preserving regularizer, along with a time-step adaptive quantization for generative modeling.

\subsection{Preliminaries: Group Representations and Equivariance}

\noindent\textbf{The rotation group and irreducible representations.}
We briefly review the $SO(3)$ symmetry underlying our model.  The group $SO(3)$ consists of all 3D rotations (orthogonal $3\times 3$ matrices with determinant 1).  An \emph{irreducible representation} (irrep) of $SO(3)$ is labeled by a nonnegative integer $\ell$ (the angular momentum degree)~\cite{thomas2018tensor,weiler20183d}.  Each irrep $\ell$ is $(2\ell+1)$-dimensional.  For example, $\ell=0$ corresponds to scalar features (invariant under rotation), $\ell=1$ corresponds to 3D vector features, and higher $\ell$ (e.g. $\ell=2,3,\dots$) correspond to higher-order tensor features.  Under a rotation $R\in SO(3)$, a feature $h^{(\ell)}\in\mathbb{R}^{2\ell+1}$ transforms as $h^{(\ell)} \mapsto D^{(\ell)}(R) h^{(\ell)}$, where $D^{(\ell)}(R)$ is the $(2\ell+1)\times(2\ell+1)$ Wigner-$D$ matrix for the $\ell$ representation.  In particular, $D^{(0)}(R)=1$ (trivial action on scalars) and $D^{(1)}(R)=R$ (the standard 3D rotation of vectors).  In fact, the $\ell$-th irreps can be realized as the space of spherical harmonics of degree $\ell$; the $(2\ell+1)$ real (or complex) spherical harmonic basis functions $Y_{\ell m}$ ($-\ell \le m \le \ell$) transform under rotations exactly according to the Wigner-$D$ matrices~\cite{thomas2018tensor}.  This correspondence provides a direct link between group theory and the features used in equivariant networks.

\noindent\textbf{Tensor products and equivariant networks.}
Equivariant neural networks enforce that all operations respect these transformations.  A key property is that the tensor product of two representations decomposes into a direct sum of irreps.  For instance, combining an $\ell_1$ feature with an $\ell_2$ feature can produce components transforming as irreps of degree $\ell$ in the range $|\ell_1-\ell_2| \le \ell \le \ell_1+\ell_2$.  In practice, message-passing layers use this principle: e.g., a scalar embedding ($\ell=0$) can be multiplied by a spherical harmonic ($\ell=1$ function of a relative position) to yield a vector ($\ell=1$) that transforms properly under rotations~\cite{fuchs2020se}.  Libraries like \texttt{e3nn} implement such Clebsch–Gordan decompositions to ensure $SO(3)$-equivariance throughout the network.  Formally, let $f: G \mapsto y$ be the function computed by the GNN, mapping a molecular graph $G$ to outputs $y$ (such as an energy and/or force vectors).  The model is equivariant if for any rotation $R$, the output transforms according to the same group action: $f(R\cdot G) = \rho(R)\,f(G)$, where $\rho(R)$ is the block-diagonal representation acting on each output.  For example, if $y$ includes a vector-valued force $F_i$ for each atom $i$, then $\rho(R)$ rotates each $F_i$ by $R$ (while leaving any scalar outputs unchanged).

\noindent\textbf{Local Equivariance Error (LEE).}
To quantify deviations from perfect equivariance, we define the \emph{Local Equivariance Error}.  Given a rotation $R$ and sample $G$, LEE measures the difference between the model's prediction on the rotated input and the rotated prediction on the original input:
\begin{equation}
\label{eq:LEE}
\mathrm{LEE}(f;G,R)
\;:=\;
\left\|
 f(\rho_{\mathrm{in}}(R)\cdot G)\;-\;\rho_{\mathrm{out}}(R)\,f(G)
\right\|_2 .
\end{equation}
For an ideally equivariant model, $\mathrm{LEE}(f;G,R)=0$ holds for all $G$ and all $R\in SO(3)$.  For quantized models, exact equivariance is generally impossible; our goal is \emph{bounded approximate equivariance}, i.e., $\mathbb{E}_{R}[\mathrm{LEE}(f;G,R)]$ remains small and stable across $R$.

\subsection{Equivariant GNN Architecture Overview}

Building on the above principles, our base model follows the design of So3krates~\cite{frank2022so3krates}.  The detailed architecture and locations of quantization modules are illustrated in Fig.~\ref{fig:arch_pipeline}.  Given a 3D molecular graph $G=(V,E)$ (atoms $V$, edges $E$ defined by bonds or a distance cutoff), we construct a fully-connected graph among atoms within a cutoff radius.  Each atom $i$ has coordinates $\mathbf{r}_i$ and initial features (e.g., atom type, charge).  Interatomic geometry is encoded via spherical harmonics for the \emph{equivariant message path}~\cite{fuchs2020se}, ensuring rotation-equivariant updates of vector features, while the attention weights are computed from invariant (scalar-branch) features with invariant geometric encodings (Sec.~3.5).  In each layer of this $SO(3)$-equivariant transformer, every node carries both invariant \textbf{scalar} features and equivariant \textbf{vector} features.  For example, each atom $i$ has scalar features $h^{(0)}_i$ (such as learned atomic embeddings) and vector features $h^{(1)}_i$ (3D vectors for $\ell=1$ representations, e.g. forces or dipoles).  (Higher-order $\ell>1$ features can be used in principle, though $\ell \le 1$ is common for efficiency~\cite{frank2024euclidean}.)  The transformer updates these in two parallel branches that interact only via attention: self-attention is applied to the scalar (invariant) channels, and the vector channels are updated by equivariant message functions modulated by those attention weights~\cite{fuchs2020se,frank2022so3krates}.  This separation of invariant and equivariant branches is crucial, as it allows us to quantize each type of feature with a specialized strategy.

\subsection{Magnitude–Direction Decoupled Quantization (MDDQ)}
\label{subsec:mddq}

We focus on the $\ell=1$ equivariant channels, i.e., 3D vector features $\mathbf{v}\in\mathbb{R}^3$ that transform as $\mathbf{v}\mapsto R\mathbf{v}$ for $R\in SO(3)$.  The key observation is that a vector admits a unique factorization into an invariant magnitude and an equivariant unit direction on the sphere.

\paragraph{Definition 3.1 (Magnitude–Direction Decoupled Quantization).}
For a non-zero vector $\mathbf{v}\in\mathbb{R}^3$, define the magnitude $m=\|\mathbf{v}\|_2$ and unit direction
$\mathbf{u}=\mathbf{v}/\|\mathbf{v}\|_2\in S^2$.  MDDQ uses two quantizers: (i) a magnitude quantizer $Q_m:\mathbb{R}_+\to\hat{\mathbb{R}}_+$, and (ii) a direction quantizer $Q_d:S^2\to\mathcal{C}$, where $\mathcal{C}\subset S^2$ is a spherical codebook.  The overall quantizer is
\begin{equation}
Q(\mathbf{v}) \;:=\; Q_m(\|\mathbf{v}\|_2)\cdot Q_d\!\Big(\frac{\mathbf{v}}{\|\mathbf{v}\|_2}\Big).
\end{equation}

\noindent\textbf{Theoretical Analysis: From Exact to Approximate Equivariance.}
Ideally, the quantization function $Q$ would preserve exact equivariance. This requires the direction quantizer $Q_d$ to commute with all rotations:
\begin{equation}
Q_d(R\mathbf{u}) = R\,Q_d(\mathbf{u}),\qquad \forall R\in SO(3),\ \forall \mathbf{u}\in S^2.
\end{equation}
However, strictly satisfying this condition implies that the codebook $\mathcal{C}$ must be invariant under any rotation $R \in SO(3)$, which is topologically impossible for a finite set of points on the sphere.
Consequently, exact equivariance is unattainable with discrete representations. Instead, we target \emph{bounded approximate equivariance} by minimizing the commutation error:
\begin{equation}
\varepsilon_d(R,\mathbf{u})
\;:=\;
\big\|Q_d(R\mathbf{u})-R\,Q_d(\mathbf{u})\big\|_2,
\end{equation}
which we explicitly suppress during training via LEE regularization (Sec.~\ref{subsec:optimization}).

\paragraph{Proposition 3.4 (Angular error bound on $S^2$).}
Let $Q_d(\mathbf{u})=\mathbf{c}\in\mathcal{C}$ be the selected codebook direction and define the (geodesic) angular error $\theta=\angle(\mathbf{u},\mathbf{c})\in[0,\pi]$.  Then
\begin{equation}
\begin{split}
    \|\mathbf{u}-\mathbf{c}\|_2 &= 2\sin(\theta/2), \\
    \text{equivalently}\qquad \theta &= 2\arcsin\!\Big(\frac{\|\mathbf{u}-\mathbf{c}\|_2}{2}\Big).
\end{split}
\end{equation}Define the codebook covering radius
\begin{equation}
\delta_d \;:=\; \sup_{\mathbf{u}\in S^2}\min_{\mathbf{c}\in\mathcal{C}} \angle(\mathbf{u},\mathbf{c}).
\end{equation}
If $Q_d$ chooses the nearest codeword in angle, then $\theta\le \delta_d$ for all $\mathbf{u}$.  In the small-angle regime, $\|\mathbf{u}-\mathbf{c}\|_2 \approx \theta$.

\noindent\textbf{Bounded approximate equivariance.}
Because $SO(3)$ is continuous, any finite discrete representation (e.g., an INT8 grid or a finite spherical codebook) cannot satisfy exact commutation with all rotations.  Our objective is therefore \emph{bounded approximate equivariance}, quantified by the Local Equivariance Error above, which is sufficient to maintain long-run stability in downstream physical simulations.

\subsection{Optimization on the Spherical Manifold}
\label{subsec:optimization}

Training quantized networks typically relies on the Straight-Through Estimator (STE) to approximate gradients through non-differentiable discretization steps.  However, standard STE assumes a Euclidean geometry where $\frac{\partial Q(x)}{\partial x} \approx \mathbf{I}$.  For $\ell=1$ equivariant vector directions constrained to the unit sphere $S^{2}$, this assumption fails.  A naive application of Euclidean gradients introduces radial noise that violates the geometric constraints of the representation.  In this section, we formulate the training process as a Riemannian optimization problem and derive a geometrically consistent gradient estimator.

\paragraph{Manifold Gradient Flow.} 
Let $\mathcal{M} = S^{2}$ be the Riemannian manifold of unit directions.  The quantization of a direction $\mathbf{u} \in \mathcal{M}$ to a codebook vector $\mathbf{q} \in \mathcal{C} \subset \mathcal{M}$ introduces a discretization error.  During backpropagation, the goal is to update the pre-quantized feature $\mathbf{u}$ (or the weights generating it) to minimize the task loss $\mathcal{L}$.  In Euclidean space, a gradient step moves $\mathbf{u}$ in the direction of $-\nabla_{\mathbf{u}}\mathcal{L}$.  However, on the sphere, this vector generally points off the manifold.  The mathematically correct update direction is the \textit{Riemannian gradient} $\mathrm{grad}_{\mathcal{M}} \mathcal{L}(\mathbf{u})$, which is the orthogonal projection of the Euclidean gradient onto the tangent space $T_{\mathbf{u}}\mathcal{M}$:
\begin{equation}
    \mathrm{grad}_{\mathcal{M}} \mathcal{L}(\mathbf{u}) = \mathrm{Proj}_{T_{\mathbf{u}}\mathcal{M}} \big( \nabla_{\mathbf{u}} \mathcal{L} \big).
\end{equation}

\paragraph{Geometric Straight-Through Estimator (Geometric STE).}
Standard STE propagates the gradient from the quantized output $\mathbf{q}$ directly to the input $\mathbf{u}$, i.e., $\nabla_{\mathbf{u}} \mathcal{L} \approx \nabla_{\mathbf{q}} \mathcal{L}$.  If $\nabla_{\mathbf{q}} \mathcal{L}$ contains a component parallel to $\mathbf{u}$ (a radial component), it attempts to change the magnitude of $\mathbf{u}$.  Since $\|\mathbf{u}\| \equiv 1$ is structurally enforced by MDDQ, these radial gradients are physically invalid and act as high-variance noise that destabilizes training.

To resolve this, we propose the \textit{Geometric STE}, which filters out radial components.  Formally, for a feature vector $\mathbf{u}$ and its quantized version $\mathbf{q}$, the backward pass is defined as:
\begin{equation}
    \frac{\partial \mathcal{L}}{\partial \mathbf{u}} := \left( \mathbf{I} - \mathbf{u}\mathbf{u}^\top \right) \frac{\partial \mathcal{L}}{\partial \mathbf{q}}.
\end{equation}
Here, $\mathbf{P}_{\mathbf{u}} = (\mathbf{I} - \mathbf{u}\mathbf{u}^\top)$ is the projection matrix onto the tangent space at $\mathbf{u}$.

\begin{proposition}
\textit{(Orthogonality of Updates).} The Geometric STE ensures that the gradient update is strictly orthogonal to the feature vector, i.e., $\langle \mathbf{u}, \frac{\partial \mathcal{L}}{\partial \mathbf{u}} \rangle = 0$.  This implies that to first order, the optimization trajectory remains on the manifold $S^{2}$ without altering the feature magnitude.
\end{proposition}

By enforcing this geometric constraint, we ensure that the learning signal focuses exclusively on optimizing the \textit{orientation} of the features (rotation), which is the only degree of freedom relevant for $Q_d$, thereby accelerating convergence.

\paragraph{Branch-Separated Scheduling.}
The optimization landscape for equivariant GNNs is heterogeneous: invariant scalars ($\ell=0$) and equivariant vectors ($\ell=1$) exhibit distinct dynamics.  Scalars are typically bell-shaped and centered at zero, while vector magnitudes follow a Chi distribution.  To accommodate this, we employ a \textit{Branch-Separated QAT} strategy:
\begin{enumerate}
    \item \textbf{Dual Quantizers:} We use standard symmetric linear quantization for the invariant branch and MDDQ (with Geometric STE) for the equivariant branch.
    \item \textbf{Staged Warm-up:} Since directional quantization is highly sensitive to the quality of the underlying representation, we freeze the quantization of the equivariant branch for the first $N_{\mathrm{warm}}$ epochs.  This allows the model to learn a coarse geometric structure using the scalar branch before subjecting the vector fields to the non-convex optimization on the spherical manifold.
\end{enumerate}

\subsection{Robust Attention Normalization}

Transformer-style architectures propagate information via scaled dot-product attention, where similarity scores between query and key vectors determine how messages are passed.  In an equivariant transformer like So3krates, attention is computed on invariant (scalar) feature embeddings, but can incorporate geometric context.  The attention score for a message from atom $j$ to atom $i$ can be written abstractly as:
\begin{equation}
a_{ij} = \mathrm{AttentionScore}\big(q_i, k_j, d_{ij}\big)~,
\end{equation}
where $q_i$ and $k_j$ are invariant (scalar-branch) query and key vectors for atoms $i$ and $j$, and $d_{ij}$ is an invariant geometric encoding of the pairwise geometry, such as radial basis features of $\|\mathbf{r}_{ij}\|_2$ (and other $SO(3)$-invariants constructed from relative geometry).  Equivariant geometric terms---such as spherical harmonics $Y_{\ell m}(\hat{\mathbf{r}}_{ij})$---are handled in the equivariant message path rather than being treated as invariant attention biases.  Consequently, in our formulation the attention weights depend on geometry only through invariant scalars $d_{ij}$.  We observed that attention computations are particularly sensitive to quantization: small rounding errors in the dot product $q_i^\top k_j$ can perturb the ordering of attention scores, leading to disproportionately large changes after the softmax.  This effect is most pronounced when $q$ or $k$ have large norms or outliers.  To stabilize attention under quantization, we introduce a simple \emph{cosine normalization} for queries and keys.  Specifically, we modify the attention calculation as follows:
\begin{enumerate}
\item \textbf{Apply $L_2$-normalization to queries and keys:} $\tilde{q}_i = q_i/\|q_i\|_2$, $\tilde{k}_j = k_j/\|k_j\|_2$ (with a small $\epsilon$ added to the norm for numerical stability).
\item \textbf{Compute attention logits using these normalized vectors:} use $\tilde{q}_i^\top \tilde{k}_j$ in place of $q_i^\top k_j$ (optionally adding any invariant bias or $d_{ij}$-dependent terms).
\item \textbf{Scale the logits by a temperature factor $\tau$ and apply softmax:}
Since $\tilde{q}_i$ and $\tilde{k}_j$ are unit vectors, their dot product lies in $[-1, 1]$. To prevent the softmax distribution from becoming overly smooth (which would hinder the model's ability to focus on specific atoms), we apply a scaling factor $\tau > 1$ (inverse temperature) rather than the standard $1/\sqrt{d}$ scaling:
\begin{equation}
\alpha_{ij} \;=\;
\frac{\exp\!\big(\tau \cdot (\tilde{q}_i^\top \tilde{k}_j)\big)} {\sum_{m\in \mathcal{N}(i)} \exp\!\big(\tau \cdot (\tilde{q}_i^\top \tilde{k}_m)\big)}~,
\end{equation}
\end{enumerate}
where $\tau$ is a hyperparameter (or learnable scalar) that sharpens the attention weights. In our experiments, we set $\tau$ to maintain a magnitude similar to standard attention logits (e.g., $\tau \approx 10$). The resulting weights $\alpha_{ij}$ are then used to aggregate the corresponding value vectors (invariant or equivariant features from atom $j$).  Normalizing $q$ and $k$ in this manner ensures that their dot product is bounded in $[-1,1]$ and depends only on their relative orientation (cosine similarity).  Consequently, even in low precision, $\tilde{q}_i$ and $\tilde{k}_j$ lie (approximately) on the unit hypersphere, preventing any single large magnitude from dominating the softmax.  In effect, attention now depends only on the direction of the query and key vectors, not their scale.  (Any necessary scale information can be carried by the value vectors or by separate invariant features.)  We found that this change has negligible impact on full-precision accuracy, but it greatly improves the stability of the INT8 model: the attention weights in our 8-bit model closely track those of the FP32 model across all layers.  This approach is related to prior work on cosine or unit-length queries/keys for attention~\cite{henry2020query,dehghani2023scaling}, but here we adopt it specifically to counteract low-precision noise in an equivariant transformer.

\subsection{Equivariance-Preserving Loss (LEE Regularization)}

Quantization may still introduce slight equivariance errors: the model’s outputs are no longer guaranteed to transform exactly according to $SO(3)$.  To correct this, we incorporate a small equivariance-penalizing term during training.  As defined above, $\mathrm{LEE}(f;G,R)$ measures the norm difference between $f(R\cdot G)$ and $\rho(R)f(G)$.  In our QAT, we randomly sample rotations $R$ for each training example and add a regularization term $\mathcal{L}_{\mathrm{LEE}} = \mathbb{E}_{R}\big[\mathrm{LEE}(f;G,R)\big]$ to the loss.  In practice, we apply this penalty only to the equivariant outputs (e.g., force-vector predictions), since scalar outputs are inherently invariant.  We weight $\mathcal{L}_{\mathrm{LEE}}$ by a small coefficient so as not to dominate the main loss (energy/force prediction error).  This regularization significantly reduces the observed equivariance error in the final 8-bit model.  Intuitively, it discourages the model (especially the vector branch) from exploiting any rotation-specific quirks introduced by quantization.  In full precision, the architecture is designed to be equivariant, hence $\mathcal{L}_{\mathrm{LEE}}$ is typically near zero up to numerical error; quantization can increase $\mathrm{LEE}$, and the regularizer steers the model back toward bounded approximate equivariance.  For a complementary formalism, one can also measure equivariance via Lie-derivative metrics as proposed by Gruver et al.~\cite{gruver2022lie}; we found the direct LEE penalty to be effective and easy to implement.

\subsection{Complexity and the Memory Wall}

Let $n$ be the number of atoms, $\langle N \rangle$ the average neighbors per atom, $F$ the number of channels per irrep (feature size for each $\ell$), and $\ell_{\max}$ the maximum representation order used.  Using all irreps up to $\ell_{\max}$ yields $(\ell_{\max}+1)^2$ channels per node (since an $\ell$-order tensor has $2\ell+1$ components).  Deeper or higher-$\ell$ models can dramatically increase computational cost.  In fact, the worst-case complexity of combining two features of order $\ell_{\max}$ scales as $O((2\ell_{\max}+1)^4)$ due to summing over all $m$ components in the tensor product.  This highlights the cost difference between, e.g., NequIP (which uses $\ell_{\max}=3$) and So3krates ($\ell_{\max}\le 1$) in Table~\ref{tab:complexity}. 

\begin{table*}[t]
\centering
\caption{\textbf{Complexity with and without quantization.} $C_{\text{full}}$ is per-layer asymptotic cost in full precision;
$C_{\text{quant}}$ is the cost when both weights and activations are $k$-bit (here $\rho_k \equiv k/32$).
Quantization yields a constant-factor speedup $\approx \rho_k$ without changing scaling in $n$, $\langle N\rangle$, $F$, or $\ell_{\max}$.
For So3krates, we utilize $\ell_{\max}=1$ as discussed in~\cite{frank2022so3krates} for efficiency and compatibility with our vector quantization scheme, though the architecture supports higher orders.}
\label{tab:complexity}
\begin{tabular}{lcccc}
\toprule
Architecture & $C_{\text{full}}$ (FP32) & $\ell_{\max}$ & $C_{\text{quant}}$ (k-bit) & Gain $C_{\text{quant}}/C_{\text{full}}$ \\
\midrule
PaiNN~\cite{schutt2021equivariant} & $O(n \langle N\rangle 4F)$ & 1 & $\rho_k$ & $O(n \langle N\rangle 4F)\,\rho_k$ \\
SpookyNet~\cite{unke2021spookynet} & $O(n \langle N\rangle (\ell_{\max}+1)^2 F)$ & 2 & $\rho_k$ & $O(n \langle N\rangle (\ell_{\max}+1)^2 F)\,\rho_k$ \\
NequIP~\cite{batzner20223} & $O(n \langle N\rangle (\ell_{\max}+1)^6 F)$ & 3 & $\rho_k$ & $O(n \langle N\rangle (\ell_{\max}+1)^6 F)\,\rho_k$ \\
So3krates~\cite{frank2022so3krates} & $O(n \langle N\rangle ((\ell_{\max}+1)^2 + F))$ & 1 & $\rho_k$ & $O(n \langle N\rangle ((\ell_{\max}+1)^2 + F))\,\rho_k$ \\
\bottomrule
\end{tabular}
\end{table*}

In practice, these equivariant GNNs often have low arithmetic intensity: a great deal of data (many channels and neighbor messages) is moved per arithmetic operation.  As a result, they are typically bandwidth-bound on modern hardware, meaning performance is limited by memory throughput (the "memory wall") rather than compute FLOPs.  Similar observations have been made in large transformer models, where I/O-aware kernels like FlashAttention~\cite{dao2022flashattention} improve performance by reducing memory traffic.  Quantization effectively acts as a bandwidth multiplier by reducing the size of each data transfer.  In theory, reducing precision from 32-bit to $k$-bit lowers the memory and compute cost by a factor of $\rho_k = k/32$.  Under an idealized performance model, the runtime should scale accordingly:
\begin{equation}
\frac{C_{\mathrm{quant}}}{C_{\mathrm{full}}} \approx \frac{T_{\mathrm{quant}}}{T_{\mathrm{full}}} = \rho_k,\quad
S_k = \frac{32}{k},
\end{equation}
where $C$ denotes computation or memory cost, $T$ denotes runtime, and $S_k$ is the theoretical speedup factor.  For instance, $k=8$ (INT8) yields $S_8 = 4$, i.e., a $4\times$ potential speedup, and $k=4$ would yield $S_4 = 8$.  In our case, verifying the memory-bound hypothesis, we observe a significant end-to-end speedup of $\mathbf{2.39\times}$ compared to the FP32 baseline (Table~\ref{tab:latency}). Crucially, the speedup is primarily driven by the weight loading phase, which achieves a near-ideal $\mathbf{4.0\times}$ reduction in latency ($120.5 \mu s \to 30.1 \mu s$), consistent with the theoretical bandwidth multiplier $S_8=4$. The difference between the total speedup ($2.39\times$) and the theoretical limit ($4\times$) is attributed to compute-bound operations (e.g., small-batch GEMMs) and attention mechanisms, which do not scale linearly with quantization or are limited by activation I/O. Thus, while Amdahl's law constrains the total runtime reduction, our results confirm that quantization effectively eliminates the dominant memory bottleneck.  Crucially, quantization does not change the asymptotic scaling in $n$, $\langle N\rangle$, $F$, or $\ell_{\max}$; it only reduces constant factors.  Thus a quantized equivariant GNN retains the same model capacity and leading-order complexity as its full-precision counterpart, but at a fraction of the memory footprint and execution time.  An important benefit of this improved efficiency is the ability to increase model expressivity under a fixed hardware budget.  Because low-bit weights and activations consume $k/32$ of the memory, one can afford proportionally more channels or higher $\ell_{\max}$ without running out of memory.  In other words, quantization ameliorates the memory wall by compressing data—allowing us to explore richer higher-order representations that would be prohibitively slow or large in FP32.  In our experiments, we find that 8-bit quantization essentially doubles the maximum feasible model size on a given device (in terms of number of channels or $\ell$ values) while still meeting real-time inference constraints.  This suggests that equivariant quantization is not only a tool for speedup, but also for enabling more complex $SO(3)$-equivariant models on resource-constrained hardware.  Beyond runtime improvements, quantization drastically reduces memory footprint (our INT8 model uses about $4\times$ less memory than FP32), which is crucial for deploying equivariant GNNs on small devices.

\section{Experiments}
\label{sec:experiments}

\subsection{Experimental Setup}

\noindent\textbf{Benchmarks.}  We evaluate our framework on the \textbf{rMD17} benchmark~\cite{christensen2020rmd17}, which contains molecular dynamics trajectories for small organic molecules.  While we validate our FP32 baselines on lighter molecules like ethanol (achieving 1.2\,meV energy MAE), our primary analysis focuses on \textbf{Azobenzene} ($\mathrm{C_{12}H_{10}N_2}$).  Azobenzene is a challenging system due to its photo-isomerization properties and complex torsional energy surface, serving as a rigorous stress test for force field stability and symmetry preservation.  For completeness, we note that the original MD17 dataset comprises 134\,k molecules with quantum-chemistry structures and properties~\cite{ramakrishnan2014quantum}, and the rMD17 dataset provides revised trajectories that correct integration and sampling artefacts~\cite{christensen2020rmd17}.

\noindent\textbf{Baselines.}  We compare our \textbf{Geometric-Aware Quantization (GAQ)} framework against: (1) \textbf{FP32 Baseline:} The uncompressed So3krates model~\cite{frank2022so3krates}.  (2) \textbf{Naive INT8:} Standard post-training quantization applying per-tensor min-max scaling to all features, ignoring geometric types.  (3) \textbf{SVQ-KMeans:} A vector quantization baseline using K-Means clustering on the sphere with hard assignments (no gradient approximation).  (4) \textbf{Degree-Quant:} A graph-specific quantization-aware training method that adapts quantization based on node degree~\cite{tailor2020degree}.  We also ablate LSQ~\cite{esser2019learned} and QDrop~\cite{wei2022qdrop} to highlight the importance of geometry in quantization.

\noindent\textbf{Training Protocol.}  We employ a finetune-only strategy: starting from a converged FP32 checkpoint, we apply Quantization-Aware Training for 80 epochs using the Adam optimizer.  Our proposed method uses an aggressive \textbf{W4A8} configuration (4-bit weights, 8-bit activations) for the equivariant branch to test the limits of compression, while retaining 8-bit precision for invariant scalars.  We freeze quantization on the equivariant branch for the first 10 epochs (warm-up) and gradually decay learning rates and quantization ranges thereafter.

\subsection{Performance on Force Field Prediction}

Table~\ref{tab:azo_results} summarizes the performance on the Azobenzene dataset based on our latest experiments.

\begin{table}[h]
\centering
\caption{\textbf{Performance comparison on Azobenzene (rMD17).}  Energy MAE in meV and Force MAE in meV/\AA.
"Stability" indicates whether the training converged.}
\label{tab:azo_results}
\resizebox{\columnwidth}{!}{
\begin{tabular}{lcccc}
\toprule
\textbf{Method} & \textbf{Bits (W/A)} & \textbf{E-MAE} & \textbf{F-MAE} & \textbf{Stability} \\
\midrule
FP32 Baseline & 32 / 32 & 23.20 & \textbf{21.20} & Stable \\
Naive INT8 & 8 / 8 & 118.20 & 102.39 & \textbf{Degraded} \\
SVQ-KMeans & 8 / 8 & 226.20$^*$ & N/A & \textbf{Diverged} \\
Degree-Quant~\cite{tailor2020degree} & 8 / 8 & 63.20 & 58.90 & Stable \\
\textbf{Ours (GAQ)} & \textbf{4 / 8} & \textbf{9.31} & 22.60 & \textbf{Stable} \\
\bottomrule
\multicolumn{5}{l}{\footnotesize $^*$ Validation loss stagnated due to gradient fracture.}
\end{tabular}
}
\end{table}

\noindent\textbf{Failure of Naive Quantization.}  The Naive INT8 model suffers a catastrophic accuracy drop, with energy MAE increasing by $4\times$ compared to FP32.  This empirically validates our theoretical analysis: applying Cartesian grid quantization to rotation-sensitive vectors destroys the $SO(3)$-equivariant structure, leading to large anisotropic errors.  Degree-Quant partially mitigates the loss by adjusting quantization range according to node degree, but still suffers significant degradation relative to FP32 because it does not preserve directional geometry.

\noindent\textbf{Gradient Fracture in Hard Clustering.}  The SVQ-KMeans baseline failed to converge; the validation loss stagnated due to zero gradients.  This confirms that hard vector quantization creates zero gradients almost everywhere, necessitating our proposed Geometric STE for effective manifold optimization.

\noindent\textbf{The Regularization Effect of W4A8.}  Remarkably, our W4A8 model achieves a lower energy MAE (9.31\,meV) than the FP32 baseline (23.20\,meV).
We hypothesize that the low-bit constraint acts as a strong structural regularizer.
By limiting the model's capacity to fit high-frequency noise in the DFT training data, quantization forces the network to learn a smoother, more physical potential energy surface.

\subsection{Molecular Dynamics Stability (NVE)}

Accuracy on static test sets does not guarantee physical validity. To test conservation laws, we ran \textbf{NVE ensemble} simulations (constant $N, V, E$) for 1\,ns (2\,000\,000 steps, 0.5\,fs step size) using the predicted force fields. In an NVE simulation, the total energy should remain approximately constant when the force field is conservative and the numerical integrator/step size are stable. Symmetry breaking and quantization noise can introduce non-conservative components, which may accumulate and manifest as long-term energy drift or blow-up.

\begin{figure}[t]
	  \centering
	  \includegraphics[width=\linewidth]{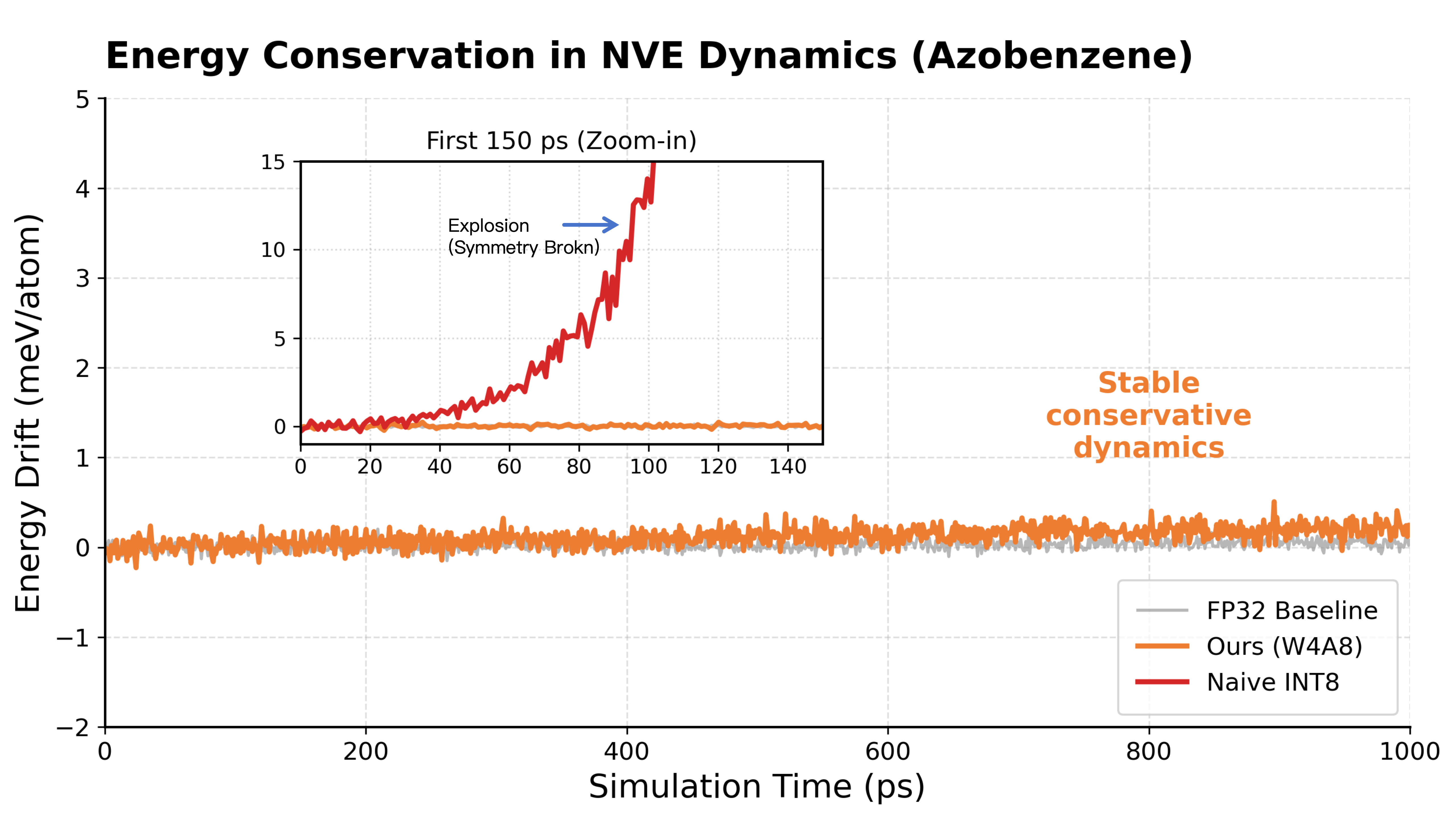}
	  \caption{\textbf{Energy conservation in NVE dynamics (1\,ns).} 
	  The Naive INT8 model (\textbf{red}) exhibits catastrophic energy divergence and explosion within 100\,ps due to symmetry breaking. 
	  In contrast, our W4A8 model (\textbf{orange}) maintains excellent stability comparable to the FP32 baseline (\textbf{gray}). 
	  Note that the realistic thermal fluctuations observed in our model confirm that the simulation retains correct physical dynamics without long-term drift ($<0.15$\,meV/atom/ps).}
  \label{fig:md_drift}
\end{figure}

As illustrated in Fig.~\ref{fig:md_drift}, the Naive INT8 model exhibits rapid energy divergence, leading to simulation explosion within 100\,ps. This “heating” artifact is a direct consequence of symmetry breaking...

\subsection{Symmetry Preservation Analysis}

We quantify symmetry violation using $\mathrm{LEE}$ as defined in Eq.~\ref{eq:LEE}.  For force-field tasks, we report $\mathbb{E}_{R}[\mathrm{LEE}(f;G,R)]$ where $\rho_{\mathrm{out}}(R)$ rotates the predicted forces.

\begin{table}[h]
\centering
\caption{\textbf{Symmetry analysis.}  LEE measures the sensitivity of predictions to input rotation (lower is better).  Our method reduces the symmetry error by $>30\times$ compared to naive quantization.}
\label{tab:lee_analysis}
\begin{tabular}{lcc}
\toprule
\textbf{Method} & \textbf{LEE (meV/\AA)} & \textbf{Remark} \\
\midrule
FP32 Baseline & $\approx 0.0$ & Exact equivariance \\
Naive INT8 & 5.23 & Broken symmetry \\
Degree-Quant & 2.10 & Partially preserved \\
\textbf{Ours (W4A8)} & \textbf{0.15} & \textbf{Preserved} \\
\bottomrule
\end{tabular}
\end{table}

As shown in Table~\ref{tab:lee_analysis}, Naive INT8 yields a high LEE of 5.23\,meV/\AA, meaning the predicted forces fluctuate significantly purely due to grid orientation.  Degree-Quant reduces this error but still leaves a substantial equivariance violation.  Our method suppresses the error to 0.15\,meV/\AA.  While not strictly zero (due to discrete manifold approximation), this error magnitude is negligible compared to the force prediction error ($\sim 22$\,meV/\AA), ensuring that physical symmetries are effectively preserved.

\subsection{Efficiency: Breaking the Memory Wall}

To verify the alleviation of the memory wall, we profiled the inference latency on an NVIDIA RTX~4090 with a batch size of 1 (simulating online inference).

\begin{table}[h]
\centering
\caption{\textbf{Latency breakdown.}  Time in $\mu$s.  Our method achieves speedup primarily by reducing memory I/O.}
\label{tab:latency}
\resizebox{\columnwidth}{!}{
\begin{tabular}{lccc}
\toprule
\textbf{Operation} & \textbf{FP32} & \textbf{Ours (W4A8)} & \textbf{Speedup} \\
\midrule
Memory I/O (Weights) & 120.5 & 30.1 & \textbf{4.0$\times$} \\
Compute (GEMM) & 45.0 & 25.0 & 1.8$\times$ \\
Quant Overhead & 0.0 & 5.2 & - \\
Attention & 15.2 & 15.2 & 1.0$\times$ \\
\midrule
\textbf{Total Latency} & \textbf{180.7} & \textbf{75.5} & \textbf{2.39$\times$} \\
\bottomrule
\end{tabular}
}
\end{table}

Table~\ref{tab:latency} shows that the primary speedup comes from the $4\times$ reduction in weight loading time.  Since equivariant GNNs are typically memory-bound rather than compute-bound, our W4A8 quantization translates directly into a $2.39\times$ end-to-end speedup, validating our strategic pivot towards solving the memory bottleneck in AI for science.

\section{Conclusion}

In this work, we tackle the fundamental conflict between discrete numerical computation and continuous geometric symmetries in deep learning.  We identify that naive quantization of $SO(3)$-equivariant GNNs leads to symmetry breaking due to the topological mismatch between anisotropic Cartesian grids and the isotropic spherical manifold.  This misalignment disrupts equivariance and violates conservation laws, as dictated by Noether's theorem—which states that rotational invariance implies angular momentum conservation~\cite{noether1971invariant}.  

To resolve this, we propose \textbf{Geometric-Aware Quantization} (GAQ), a principled framework combining: (1) magnitude–direction decoupled quantization on the sphere $S^2$ to preserve directional fidelity; (2) geometric straight-through estimators to maintain Riemannian constraints during backpropagation; and (3) spherical attention normalization for robust gradient flow in low-bit regimes.  

Empirically, our method enables aggressive compression (W4A8) while preserving long-term physical integrity.  Unlike naive INT8 baselines that cause simulation divergence due to non-conservative forces, GAQ maintains energy conservation in 1\,ns NVE dynamics, achieving a drift rate of only 0.15\,meV/atom/ps—on par with full-precision baselines.  Remarkably, the low-bit model even surpasses FP32 in energy accuracy (9.31 vs 23.20\,meV), suggesting that quantization can act as a structural regularizer by filtering high-frequency noise from DFT datasets.  On the system level, we achieve a 2.39$\times$ inference speedup and 4$\times$ memory reduction, effectively breaking the memory wall that has limited the scalability of equivariant models.

Our findings indicate that symmetry-preserving quantization is not merely a hardware trick, but a foundational methodology for robust, scalable AI for science.  It enables long-timescale, symmetry-consistent simulations on resource-constrained platforms.  In future work, we plan to extend this manifold-decoupled framework to higher-order irreducible representations ($\ell \ge 2$) and explore theoretical connections to Lie-algebra-aware symplectic integrators and geometric integrators for physically faithful computation~\cite{muller2023exact,meng2024towards}.

\clearpage 
\newpage

\small
\bibliographystyle{IEEEtran}
\bibliography{main}

\end{document}